%% file: main.tex
\pdfoutput=1

\documentclass[11pt]{article}
\usepackage[dvipsnames]{xcolor}
\usepackage[final]{emnlp2021}
\usepackage{times}
\usepackage{latexsym}

\usepackage{MnSymbol}

\usepackage{microtype}
\usepackage[utf8]{inputenc}
\usepackage{graphicx}
\usepackage{caption}
\usepackage{subcaption}
\usepackage{booktabs}
\usepackage{multirow}
\usepackage{amsmath}
\usepackage{bbold}
\usepackage{algorithm}
\usepackage{algorithmic}
\usepackage[export]{adjustbox}

\usepackage[capitalise,noabbrev]{cleveref}

\usepackage{macros}

\title{Are Gender-Neutral Queries Really Gender-Neutral?\\ Mitigating Gender Bias in Image Search}

\author{Jialu Wang, Yang Liu, Xin Eric Wang \\
Department of Computer Science and Engineering \\
  University of California, Santa Cruz \\
  \texttt{\{faldict,yangliu,xwang366\}@ucsc.edu}}

\date{}

\begin{document}
\maketitle
\begin{abstract}
Internet search affects people's cognition of the world, so mitigating biases in search results and learning fair models is imperative for social good. We study a unique gender bias in image search in this work: the search images are often gender-imbalanced for gender-neutral natural language queries. We diagnose two typical image search models, the specialized model trained on in-domain datasets and the generalized representation model pre-trained on massive image and text data across the internet. Both models suffer from severe gender bias. Therefore, we introduce two novel debiasing approaches: an in-processing fair sampling method to address the gender imbalance issue for training models, and a post-processing feature clipping method base on mutual information to debias multimodal representations of pre-trained models. Extensive experiments on MS-COCO~\cite{Lin2014MicrosoftCC} and Flickr30K~\cite{Flickr30K} benchmarks show that our methods significantly reduce the gender bias in image search models.
\end{abstract}

\input{sections/introduction}
\input{sections/metric}
\input{sections/approach}
\input{sections/experiment}
\input{sections/related-work}
\input{sections/conclusion}

\section*{Broader Impact}
The algorithmic processes behind modern search engines, with extensive use of machine learning algorithms, have great power to determine users' access to information~\cite{764d88eb1b1043249afc97900350d6a6}. Our research provides evidence that unintentionally using image search models trained either on in-domain image retrieval data sets or massive corpora across the internet may lead to unequal inclusiveness between males and females in image search results, even when the search terms are gender-neutral. This inequity can and do have significant impacts on shaping and exaggerating gender stereotype in people's minds~\cite{gender-image-search}.

This work offers new methods for mitigating gender bias in multimodal models, and we regard the algorithms proposed in this paper have the potentials to be deployed in real-world systems. We conjecture that our methods may contribute to driving the development of responsible image search engines with other fairness issues. For instance, we would encourage future works to understand and mitigate the risks arising from other social biases, like racial bias, in image search results. We would also encourage researchers to explore whether the methodology presented in this work could be generalized to quantify and mitigate other bias measures. 

Our work has limitations. The gender bias measures and the debiasing methods proposed in this study require acquiring the gender labels of images. Our method for identifying the gender attributes of people portrayed in the images is limited: we make use of the contextual cues in the human-annotated captions from the image datasets. The accuracy of such a proxy-based method heavily relies on the coverage of gendered nouns and the inclusiveness of gendered language in the original human annotations. The corruption of gender labels, due to missing gendered words or inappropriate text pre-processing steps, may introduce biases we have not foreseen into the evaluated metrics. Additionally, the gendered word lists are collected from English corpora and may differ in other languages or cultures. It is possible that blind application of our methods by improperly acquiring the gender labels may create image search models that produce even greater inequality, which is very much discouraged. This limitation arises from the unavailability of such sensitive attributes in the source datasets. The lack of relevant data for studying gender bias in image search, and the concerns about how to acquire the gender attributes while preserving the privacy of people concerned, is itself an important question in this area. We believe this research would benefit when richer datasets become available.

\section*{Acknowledgements}
The authors would like to thank anonymous reviewers for their constructive comments. This work is supported by the UC Santa Cruz Startup Funding, and the National Science Foundation (NSF) under grants IIS-2040800 and CCF-2023495.

\bibliography{references}
\bibliographystyle{acl_natbib}

\appendix
\input{sections/appendix}

\end{document}

%% file: sections/introduction.tex
\section{Introduction}\label{sec:introduction}
Internet information is shaping people's minds. The algorithmic processes behind modern search engines, with extensive use of machine learning, have great power to determine users' access to information~\cite{764d88eb1b1043249afc97900350d6a6}. These information systems are biased when results are systematically slanted in unfair discrimination against protected groups~\cite{BiasInComputerSystems}.

\begin{figure*}
\begin{center}
    \begin{subfigure}[b]{0.19\linewidth}
    \centering
    \includegraphics[width=\linewidth, height=\linewidth]{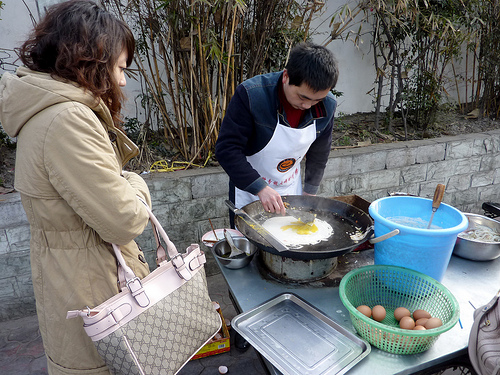}
    \end{subfigure}
    \begin{subfigure}[b]{0.19\linewidth}
    \centering
    \includegraphics[width=\linewidth,height=\linewidth]{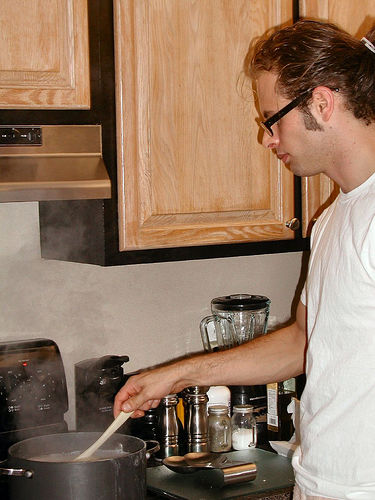}
    \end{subfigure}
    \begin{subfigure}[b]{0.19\linewidth}
    \centering
    \includegraphics[width=\linewidth,height=\linewidth]{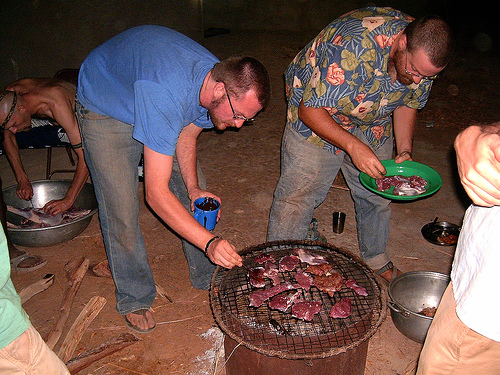}
    \end{subfigure}
    \begin{subfigure}[b]{0.19\linewidth}
    \centering
    \includegraphics[width=\linewidth,height=\linewidth]{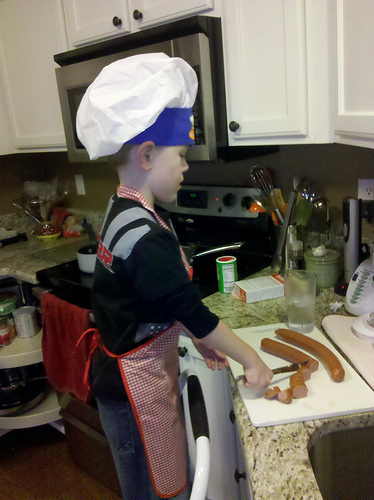}
    \end{subfigure}
    \begin{subfigure}[b]{0.19\linewidth}
    \centering
    \includegraphics[width=\linewidth,height=\linewidth]{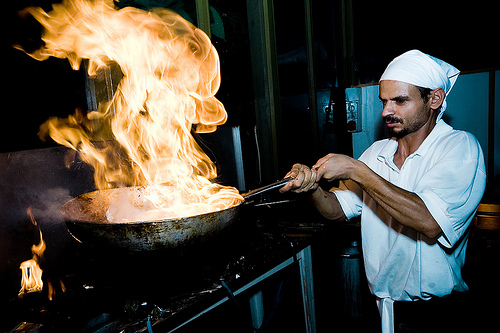}
    \end{subfigure}
    \vskip0.15\baselineskip
    \begin{subfigure}[b]{0.19\linewidth}
    \centering
    \includegraphics[width=\linewidth, height=\linewidth]{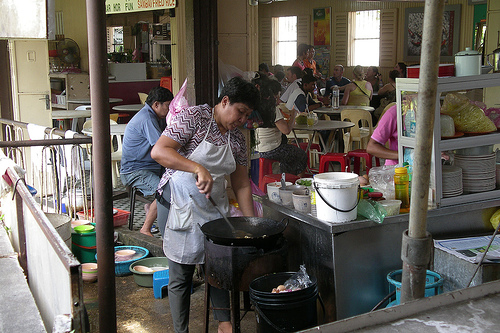}
    \end{subfigure}
    \begin{subfigure}[b]{0.19\linewidth}
    \centering
    \includegraphics[width=\linewidth,height=\linewidth]{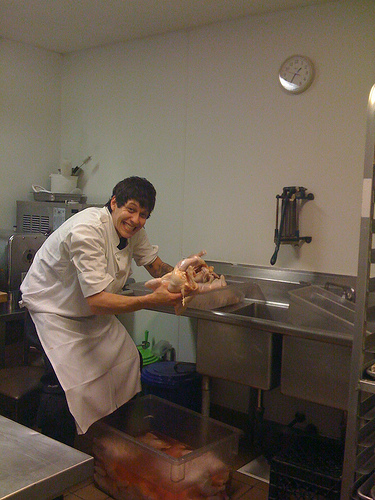}
    \end{subfigure}
    \begin{subfigure}[b]{0.19\linewidth}
    \centering
    \includegraphics[width=\linewidth,height=\linewidth]{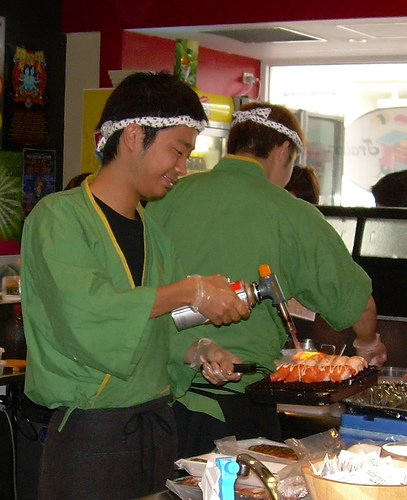}
    \end{subfigure}
    \begin{subfigure}[b]{0.19\linewidth}
    \centering
    \includegraphics[width=\linewidth,height=\linewidth]{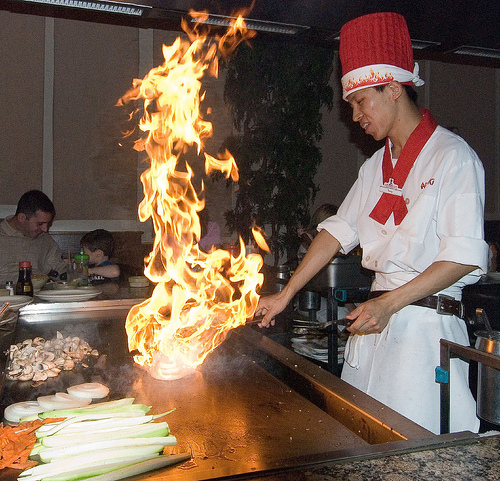}
    \end{subfigure}
    \begin{subfigure}[b]{0.19\linewidth}
    \centering
    \includegraphics[width=\linewidth,height=\linewidth]{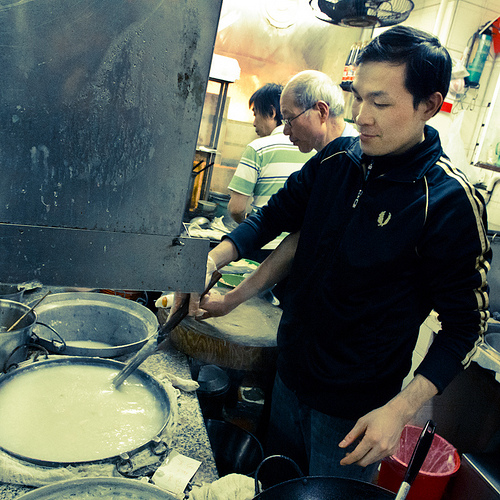}
    \end{subfigure}
\end{center}
    \caption{Gender bias in image search. We show the top-10 retrieved images for searching ``a person is cooking'' on the Flickr30K~\cite{Flickr30K} test set using a state-of-the-art model~\cite{CLIP}. Despite the gender-neutral query, only 2 out of 10 images are depicting female cooking.}
    \label{fig:case_study}
\end{figure*}

Gender bias is a severe fairness issue in image search. \cref{fig:case_study} shows an example: given a gender-neutral natural language query ``a person is cooking", only 2 out of 10 images retrieved by an image search model~\cite{CLIP} depict females, while equalized exposure for male and female is expected. Such gender-biased search results are harmful to society as they change people's cognition and worsen gender stereotypes~\cite{gender-image-search}. Mitigating gender bias in image search is imperative for social good.

In this paper, we formally develop a framework for quantifying gender bias in image search results, where text queries in English\footnote{This study is conducted on English corpora. We will assume the text queries are all English queries hereafter.} are made gender-neutral, and gender-balanced search images are expected for models to retrieve. To evaluate model fairness, we use the normalized difference between masculine and feminine images in the retrieved results to represent gender bias. We diagnose the gender bias of two primary families of multimodal models for image search: (1) the specialized models that are often trained on in-domain datasets to perform text-image retrieval, and (2) the general-purpose representation models that are pre-trained on massive image and text data available online and can be applied to image search. Our analysis on MS-COCO~\cite{Lin2014MicrosoftCC} and Flickr30K~\cite{Flickr30K} datasets reveals that both types of models lead to serious gender bias issues (e.g., nearly 70\% of the retrieved images are masculine images).

To mitigate gender bias in image search, we propose two novel debiasing solutions for both model families. The specialized in-domain training methods such as SCAN~\cite{SCAN} often adopt contrastive learning to enforce image-text matching by maximizing the margin between positive and negative image-text pairs. However, the gender distribution in the training data is typically imbalanced, which results in unfair model training. Thus we introduce a fair sampling (\emph{FairSample}) method to alleviate the gender imbalance during training without modifying the training data.  

Our second solution aims at debiasing the large, pre-trained multimodal representation models, which effectively learn pre-trained image and text representations to accomplish down-stream applications~\cite{NEURIPS2019_ddf35421,iGPT,SimCLR,gan2020large,chen2020uniter,CLIP}. We examine whether the representative CLIP model~\cite{CLIP} embeds human biases into multimodal representations when they are applied to the task of image search. Furthermore, we propose a novel post-processing feature clipping approach, \emph{clip}, that effectively prunes out features highly correlated with gender based on their mutual information to reduce the gender bias induced by multimodal representations. The \emph{clip} method does not require any training and is compatible with various pre-trained models.

We evaluate both debiasing approaches on MS-COCO and Flickr30K and find that, on both benchmarks, the proposed approaches significantly reduce the gender bias exhibited by SCAN and CLIP models when evaluated on the gender-neutral corpora, yielding fairer and more gender-balanced search results. In addition, we evaluate the similarity bias of the CLIP model in realistic image search results for occupations on the internet, and observe that the post-processing methods mitigate the discrepancy between gender groups by a large margin.

Our contributions are four-fold: (1) we diagnose a unique gender bias in image search, especially for gender-neutral text queries; (2) we introduce a fair sampling method to mitigate gender bias during model training; (3) we also propose a novel post-processing clip method to debias pre-trained multimodal representation models; (4) we conduct extensive experiments to analyze the prevalent bias in existing models and demonstrate the effectiveness of our debiasing methods.

%% file: sections/metric.tex
\section{Gender Bias in Image Search}
In an image search system, text queries may be either gender-neutral or gender-specific. Intuitively, when we search for a gender-neutral query like ``a person is cooking'', we expect a fair model returning approximately equal proportions of images depicting men and women. For gender-specific queries, an unbiased image search system is supposed to exclude images with misspecified gender information. This intention aligns with seeking more accurate search results and would be much different from the scope of measuring gender bias in gender-neutral cases. Therefore, we focus on identifying and quantifying gender bias when only searching for gender-neutral text queries.

\subsection{Problem Statement}\label{sec:problem-statement}
Given a text query provided by the users, the goal of an image search system is to retrieve the matching images from the curated images. In the domain of multi-modality, given the dataset $\{(v_n,c_n)\}_{n=1}^N$ with $N$ image-text pairs, the task of image search aims at matching every image $v$ based on the providing text $c$. We use $\imageset =\{v_n\}_{n=1}^N$ to denote the image set and $\textset = \{c_n\}_{n=1}^N$ to denote the text set. Given a text query $c \in \textset$ and an image $v \in \imageset$, image retrieval models often predict the similarity score $S(v, c)$ between the image and text. One general solution is to embed the image and text into a high-dimensional representation space and compute a proper distance metric, such as Euclidean distance or cosine similarity, between vectors~\cite{Wang2014LearningFI}. We take cosine similarity for an example:
\begin{equation}\label{eq:similarity}
\begin{split}
    & S(v, c) = \frac{\vec{v} \cdot \vec{c}}{\|\vec{v}\| \|\vec{c}\|} \\
    \textnormal{s.t.}\ &\ \vec{v} = \mathsf{image\_encoder}(v) \\
    &\ \vec{c} = \mathsf{text\_encoder}(c)
\end{split}
\end{equation}
The image search system outputs a set of top-$K$ retrieved images $\mathcal{R}_{K}(c)$ with the highest similarity scores. In this work, we assume that when evaluating on test data, $\forall c \in \mathcal{C}$, the text query $c$ is written in gender-neutral language.

\subsection{Measuring Gender Bias}\label{sec:measure-gender-bias}
The situations of image search results are complex: there might be no people, one person, or more than one person in the images. Let $g(v) \in \{\male, \female, \neutral\}$ represent the gender attribute of an image $v$. Note that in this study gender refers to biological sex~\citealp{larson-2017-gender}. We use the following rules to determine $g(v)$: $g(v) = \male$ when there are only men in the image, $g(v) = \female$ when there are only women in the image, otherwise $g(v) = \neutral$. 

Portraits in image search results with different gender attributes often receive unequal exposure. Inspired by ~\citet{gender-image-search} and~\citet{Zhao2017MenAL}, we measure gender bias in image search by comparing the proportions of masculine and feminine images in search results. Given the set of retrieved images $\mathcal{R}_{K}(c)$, we count the images depicting males and females
\begin{align*}
    N_\male & = \sum_{v \in \mathcal{R}_K(c)} \Indicator[g(v) = \male], \\
    N_\female & = \sum_{v \in \mathcal{R}_K(c)} \Indicator[g(v) = \female],
\end{align*}
and define the gender bias metric as:
\begin{equation}
    \Delta_K(c) = \begin{cases} 
    0, & \text{ if $N_\male + N_\female = 0$ } \\
    \frac{N_\male - N_\female}{N_\male + N_\female}, & \text{ otherwise} \end{cases}
\end{equation}
We don't take absolute values for measuring the direction of skewness, i.e., if $\Delta_K(c) > 0$ it skews towards males. Note that a similar definition of gender bias $\frac{N_\male}{N_\male + N_\female}$ in \citet{Zhao2017MenAL} is equivalent to $(1 + \Delta(c))/2$. But our definition of gender bias considers the special case when none of the retrieved images are gender-specific, i.e., $N_\male + N_\female = 0$. For the whole test set, we measure the mean difference over all the text queries:
\begin{equation}\label{eq:fairness}
\bias@ K = \frac{1}{|\textset|} \sum_{c \in \textset} \Delta_K(c)
\end{equation}

%% file: sections/approach.tex
\section{Mitigating Gender Bias in Image Search}\label{sec:methodology}
There are two fashions of multimodal models for the image search task. One is to build a specialized model that could embed image and text into representation vectors with measurable similarity scores. The other is to use general-purpose image-text representations pre-trained on sufficiently big data and compute a particular distance metric. We focus on two representative models, SCAN~\cite{SCAN} and CLIP~\cite{CLIP}, for both fashions. For the first fashion, we propose an in-processing learning approach to ameliorate the unfairness caused by imbalanced gender distribution in training examples. This approach builds on contrastive learning but extends with a \textit{fair sampling} step. The in-processing solution requires full training on in-domain data examples. For the second fashion, we propose a post-processing feature \textit{clipping} technique to mitigate bias from an information-theoretical perspective. This approach is compatible with pre-trained models and is light to implement without repeating training steps. 

\subsection{In-processing Debiasing: Fair Sampling}\label{sec:fairsample}
Image search models in the first fashion are often trained under the contrastive learning framework~\cite{contrastive-learning-review}. For our in-processing debiasing approach, we now explain the two primary components, contrastive learning and fair sampling, within our context. 

\paragraph{Contrastive Learning}
We start by formally introducing the standard contrastive learning framework commonly used in previous works~\cite{SCAN,Chen2020AdaptiveOQ} for image-text retrieval. Given a batch of $N$ image-text pairs $\batch = \{(v_n,c_n)\}_{n=1}^N$, the model aims to maximize the similarity scores of matched image-text pairs (positive pairs) while minimizing that of mismatched pairs (negative pairs). The representative SCAN model~\cite{SCAN}, denoted as $S(v, c)$ outputting a similarity score between image and text, is optimized with a standard hinge-based triplet loss:
\begin{align}
    \mathcal{L}_{i-t} & = \sum_{(v, c) \in \batch} [\gamma - S(v, c) + S(v, \tilde{c})]_+ \\
    \mathcal{L}_{t-i} & = \sum_{(v, c) \in \batch} [\gamma - S(v, c) + S(\tilde{v}, c)]_+
\end{align}
where $\gamma$ is the margin, $\tilde{v}$ and $\tilde{c}$ are negative examples, and $[\cdot]_+$ denotes the ramp function. $\mathcal{L}_{i-t}$ corresponds to image-to-text retrieval, while $\mathcal{L}_{t-i}$ corresponds to text-to-image retrieval (or image search). Common negative sampling strategy includes selecting all the negatives~\cite{HuangWW17}, selecting hard negatives of highest similarity scores in the mini-batch~\cite{faghri2018vse++}, and selecting hard negatives from the whole training data~\cite{Chen2020AdaptiveOQ}.
Minimizing the margin-based triplet loss will make positive image-text pairs closer to each other than other negative samples in the joint embedding space.

\paragraph{Fair Sampling}
One major issue in the contrastive learning framework is that the gender distribution in a batch of image-text pairs is typically imbalanced. Hence, the negative samples will slant towards the majority group, leading to systematic discrimination. To address this problem, we propose a fair sampling strategy. We split the batch of image-text pairs into masculine and feminine pairs based on the image's gender attribute:
\begin{align*}
    \imageset_\male & = \{v \mid g(v) = \male, (v, c) \in \batch\} \\
    \imageset_\female & = \{v \mid g(v) = \female, (v, c) \in \batch\} \\
    \imageset_\neutral & = \{v \mid g(v) = \neutral, (v, c) \in \batch\}
\end{align*}
For every positive image and text pair $(v, c) \in \batch$, we identify the gender information contained in the query $c$. If the natural language query is gender-neutral, we sample a negative image from the set of male and female images with probability $\frac{1}{2}$, respectively. Otherwise, we keep the primitive negative sampling selection strategy for keeping the model's generalization on gender-specific queries. Let $\batch^\ast$ be the batch of gender-neutral image-text pairs, the image search loss with fair sampling is:
\begin{multline}
\mathcal{L}_{t-i}^{fair} = \sum_{(v, c) \in \batch^\ast}(\frac{1}{2}\Expectation_{\bar{v} \in \imageset_\male} [\gamma - S(v, c) + S(\bar{v}, c)]_+ \\
\quad + \frac{1}{2}\Expectation_{\bar{v} \in \imageset_\female} [\gamma - S(v, c) + S(\bar{v}, c)]_+) \\
+ \sum_{(v, c) \in \batch / \batch^\ast} [\gamma - S(v, c) + S(\tilde{v}, c)]_+
\end{multline}
Empirically, we find that if we thoroughly apply the Fair Sampling strategy, the recall performance drops too much. To obtain a better tradeoff, we use a weight $\alpha$ to combine the objectives 
\[
\alpha \mathcal{L}_{t-i}^{fair} + (1-\alpha)\mathcal{L}_{t-i}
\]
as the final text-to-image loss function. We do not alter the sentence retrieval loss $\mathcal{L}_{i-t}$ during training for preserving generalization.

\subsection{Post-processing Debiasing: Feature Clipping based on Mutual Information}\label{sec:prune}
Pre-training methods have shown promising zero-shot performance on extensive NLP and computer vision benchmarks. The recently introduced CLIP model~\cite{CLIP} was pre-trained on an enormous amount of image-text pairs found across the internet to connect text and images. CLIP can encode image and text into $d$-dimensional embedding vectors, based on which we can use cosine similarity to quantify the similarity of image and text pairs. In this work, we find that the pre-trained CLIP model reaches the state-of-the-art performance but exhibits large gender bias due to training on uncurated image-text pairs collected from the internet. Although \citet{CLIP} released the pre-trained CLIP model, the training process is almost unreproducible due to limitations on computational costs and massive training data.

In order to avoid re-training of the CLIP model, we introduce a novel post-processing mechanism to mitigate the representation bias in the CLIP model. We propose to ``clip'' the dimensions of feature embeddings that are highly correlated with gender information.
This idea is motivated by the fact that an unbiased retrieve implies the independence between the covariates (active features) and sensitive attributes (gender)~\cite{barocas-hardt-narayanan}. Clipping the highly correlating covariates will return us a relatively independent and neutral set of training data that does not encode hidden gender bias.

\begin{algorithm}[tb]
\small
\caption{\textit{clip} algorithm}
\label{algo:prune}
\begin{algorithmic}
    \REQUIRE Index set $\Omega = \{1, ..., d\}$, number of clipped features $0 \leq m < d$ 
    \STATE $\mathcal{Z} \leftarrow \O$;
    \FOR {$i=1$ to $d$}
        \STATE Estimate mutual information $I(V_i; g(V))$;
    \ENDFOR
    \FOR {$j=1$ to $m$}
        \STATE $z \leftarrow \argmax \{I(V_i; g(V)): i \in \Omega / \mathcal{Z}\}$;
        \STATE $\mathcal{Z} \leftarrow \mathcal{Z} \cup \{z\}$;
    \ENDFOR
    \RETURN Index set of clipped features $\mathcal{Z}$
\end{algorithmic}
\end{algorithm}

The proposed \emph{clip} algorithm is demonstrated in \cref{algo:prune}, and we explain the key steps below. Let $\Omega = \{1,...,d\}$ be the full index set. We use $V = V_\Omega = [V_1, V_2, ..., V_d]$ to represent the variable of $d$-dimensional encoding image vectors and $g(V) \in \{\male, \female, \neutral\}$ to represent the corresponding gender attribute. The goal is to output the index set $\mathcal{Z}$ of clipped covariates that reduce the dependence between representations $V_{\Omega/\mathcal{Z}}$ and gender attributes $g(V)$. We measure the correlation between each dimension $V_i$ and gender attribute $g(V)$ by estimating their mutual information $I(V_i;g(V))$~\cite{estimate-mutual-information}:
\begin{equation}
    I(V_I;g(V)) = D_{\text{KL}}(\Probability_{(V_i,g(V))}\|\Probability_{V_i} \otimes \Probability_{g(V)})
\end{equation}
where $D_{\text{KL}}$ is the KL divergence~\cite{kullback1951}, $\Probability_{(V_i,g(V))}$ indicates the joint distribution, $\Probability_{V_i}$ and $\Probability_{g(V)}$ indicate their marginals. Next, we greedily clip $m$ covariates with highest mutual information, and construct $(d-m)$-dimensional embedding vectors $V_{\Omega / \mathcal{Z}}$. $m$ is a hyper-parameter that we will experimentally find to best trade-off accuracy and the reduced gender bias, and we show how the selection of $m$ affects the performance in \cref{sec:trade-off}. To project text representations, denoted by variable $C$, into the same embedding space, we also apply the index set $\mathcal{Z}$ to obtain clipped text embedding vectors $C_{\Omega / \mathcal{Z}}$. 

The clipped image and text representations, denoted by $\vec{v}^\ast$ and $\vec{c}^\ast$, will have a relatively low correlation with gender attributes due to the ``loss'' of mutual information. Then we compute the cosine similarity between image and text by substituting $\vec{v}^\ast$ and $\vec{c}^\ast$ into \cref{eq:similarity}:
\begin{equation}
    S(v, c) = \frac{\vec{v}^\ast \cdot \vec{c}^\ast}{\|\vec{v}^\ast\| \|\vec{c}^\ast\|} 
\end{equation}
Finally, we rank the images based on the cosine similarity between the clipped representations.

%% file: sections/experiment.tex
\begin{table*}[tb]
\small
    \centering
    \resizebox{\linewidth}{!}{
    \begin{tabular}{p{0.5\textwidth} p{0.5\textwidth}}
        \toprule
        Before Pre-processing & After Pre-processing  \\
        \midrule
        A \textcolor{orange}{\textbf{man}} with a red helmet on a small moped on a dirt road. & A \textcolor{PineGreen}{\textbf{person}} with a red helmet on a small moped on a dirt road. \\
        A little \textcolor{orange}{\textbf{girl}} is getting ready to blow out a candle on a small dessert. & A little \textcolor{PineGreen}{\textbf{child}} is getting ready to blow out a candle on a small dessert.\\
        A \textcolor{orange}{\textbf{female}} surfboarder dressed in black holding a white surfboard. & A surfboarder dressed in black holding a white surfboard. \\
        A group of young \textcolor{orange}{\textbf{men and women}} sitting at a table. & A group of young \textcolor{PineGreen}{\textbf{people}} sitting at a table. \\
        \bottomrule
    \end{tabular}}
    \caption{Samples of the constructed gender-neutral captions. For evaluation, we convert gender-specific captions to gender-neutral ones by replacing or removing the gender-specific words.}
    \label{tab:construct-neutral-query}
\end{table*}

\section{Experimental Setup}\label{sec:experiment}

\subsection{Datasets}
We evaluate our approaches on the standard MS-COCO~\cite{Chen2015MicrosoftCC} and Flickr30K~\cite{Flickr30K} datasets. Following \citet{Karpathy2017DeepVA} and \citet{faghri2018vse++}, we split MS-COCO captions dataset into 113,287 training images, 5,000 validation images and 5,000 test images.\footnote{The data is available at \url{cocodataset.org}.} Each image corresponds to 5 human-annotated captions. We report the results on the test set by averaging over five folds of 1K test images or evaluating the full 5K test images. Flickr30K consists of 31,000 images collected from Flickr.\footnote{The data is available at \url{http://bryanplummer.com/Flickr30kEntities/}.} Following the same split of \citet{Karpathy2017DeepVA,SCAN}, we select 1,000 images for validation, 1,000 images for testing, and the rest of the images for training. 

\paragraph{Identifying Gender Attributes of Images}
Sensitive attributes such as gender are often not explicitly annotated in large-scale datasets such as MS-COCO and Flickr30K, but we observe that implicit gender attributes of images can be extracted from their associated human-annotated captions. 
Therefore, we pre-define a set of masculine words and a set of feminine words.\footnote{We show the word lists in \cref{app:word-list}.}
Following~\citet{Zhao2017MenAL} and~\citet{Burns2018WomenAS} we use the ground-truth annotated captions to identify the gender attributes of images. An image will be labeled as ``male'' if at least one of its captions contains masculine words and no captions include feminine words. Similarly, an image will be labeled as ``female" if at least one of its captions contains feminine words and no captions include masculine words. Otherwise, the image will be labeled as ``gender-neutral".

\begin{figure*}[ht]
    \begin{subfigure}[b]{0.32\linewidth}
    \centering
    \includegraphics[width=\linewidth]{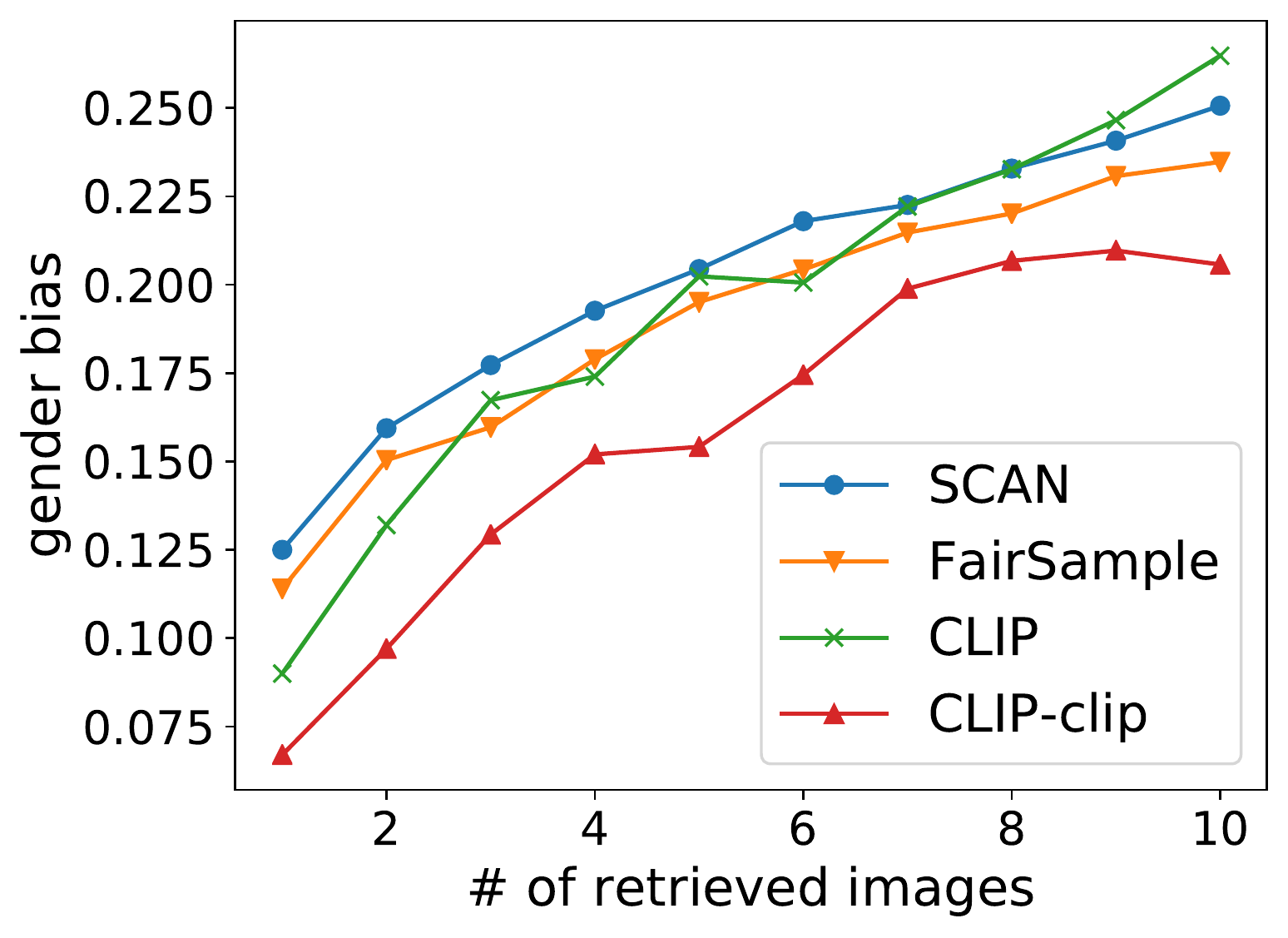}
    \caption{MS-COCO 1K Test Set.}
    \end{subfigure}
    \hfill
    \begin{subfigure}[b]{0.32\linewidth}
    \centering
    \includegraphics[width=\linewidth]{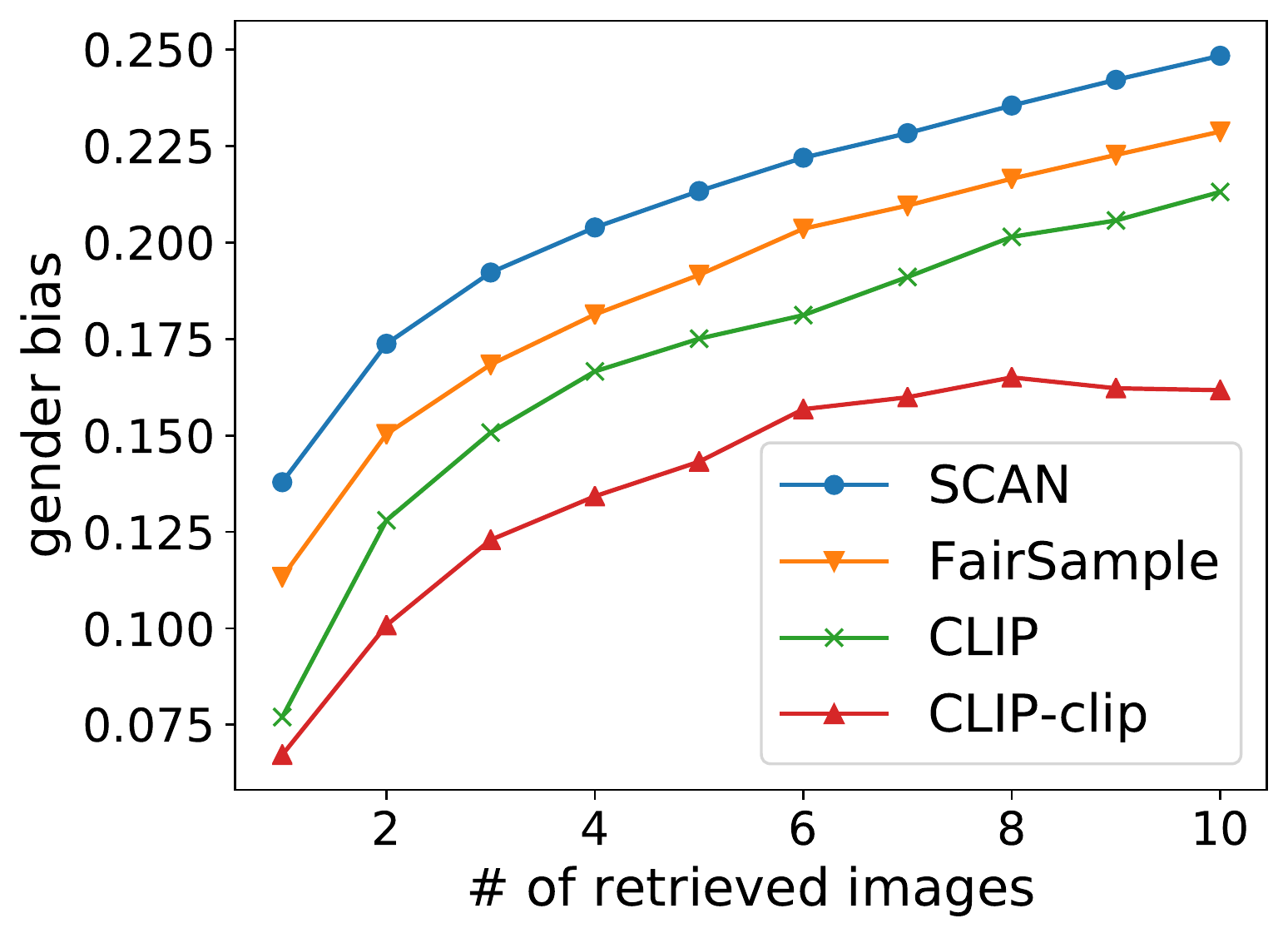}
    \caption{MS-COCO 5K Test Set.}
    \end{subfigure}
    \hfill
    \begin{subfigure}[b]{0.32\linewidth}
    \centering
    \includegraphics[width=\linewidth]{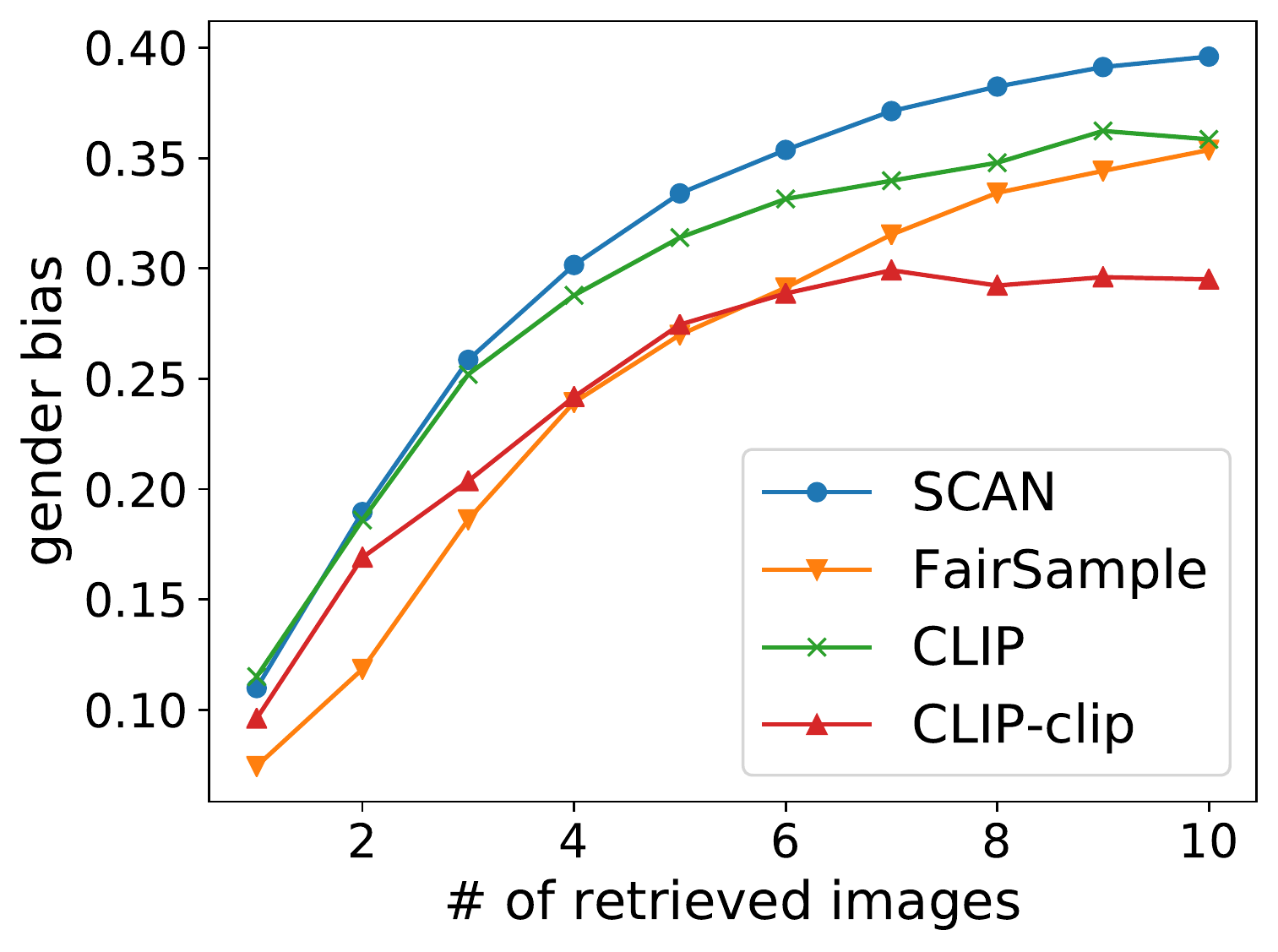}
    \caption{Flick30K Test Set.}
    \end{subfigure}
    \caption{Gender bias analysis with different top-$K$ results. }
    \label{fig:gender-bias}
\end{figure*}

\subsection{Models}
We compare the fairness performance of the following approaches:
\squishlist
\item \textbf{SCAN}~\cite{SCAN}: we use the official implementation for training and evaluation\footnote{The code is available at \url{https://github.com/kuanghuei/SCAN}.}.
\item \textbf{FairSample}: we apply the fair sampling method proposed in \cref{sec:fairsample} to the SCAN framework and adopt the same hyper-parameters suggested by \citet{SCAN} for training.
\item \textbf{CLIP}~\cite{CLIP}: we use the pre-trained CLIP model released by OpenAI.\footnote{The pre-trained model is available at \url{https://github.com/openai/CLIP}.} The model uses a Vision Transformer~\cite{dosovitskiy2021an} as the image encoder and a masked self-attention Transformer~\cite{attention} as the text encoder. The original model produces 500-dimensional image and text vectors.
\item \textbf{CLIP-clip}: we apply the feature pruning algorithm in \cref{sec:prune} to the image and text features generated by the CLIP model. We set $m=100$ and clip the image and text representations into 400-dimensional vectors.
\squishend
Note that SCAN and FairSample are trained and tested on the in-domain MS-COCO and Flickr30K datasets, while the pre-trained CLIP model is directly tested on MS-COCO and Flickr30K test sets without fine-tuning on their training sets (same for CLIP-clip as it simply drops CLIP features).

\subsection{Evaluation}
\paragraph{Gender-Neutral Text Queries}
In this study, we focus on equalizing the search results of gender-neutral text queries. In addition to the existing gender-neutral captions in the test sets, we pre-process those gender-specific captions to construct a purely gender-neutral test corpus to guarantee a fair and large-scale evaluation.
For every caption, we identify all these gender-specific words and remove or replace them with corresponding gender-neutral words. We show some pre-processing examples in \cref{tab:construct-neutral-query}.

\paragraph{Metrics} As introduced in \cref{sec:measure-gender-bias}, we employ the fairness metric in \cref{eq:fairness}, Bias@K, to measure the gender bias among the top-K images.
In addition, following standard practice, we measure the retrieval performance by Recall@K, defined as the fraction of queries for which the correct image is retrieved among the top-K images.

\section{Debiasing Results}
\subsection{Main Results on MS-COCO \& Flickr30K}
We report the results comparing our debiasing methods and the baseline methods in \cref{tab:main_result}.

\paragraph{Model Bias} 
Although the pre-trained CLIP model is evaluated without fine-tuning, we observe that it achieves a comparable recall performance with the SCAN model on MS-COCO and dominates the Flickr30K dataset. However, both models suffer from severe gender bias. Especially, the $\bias@10$ of the SCAN model on Flickr30K is $0.3960$, meaning nearly $70\%$ of the retrieved gender-specific images portray men and only $30\%$ portray women. Similarly, the CLIP model achieves $0.2648$ gender bias on MS-COCO 1K test set, indicating about $6.4$ out of 10 retrieved images portray men while about $3.6$ out of 10 portray women. Given that all of the testing text queries are gender-neutral, this result shows that severe implicit gender bias exists in image search models.

\begin{table*}[t]
\small
    \setlength{\tabcolsep}{2.5pt}
    \centering
    \begin{tabular}{l l  c c c  c c c }
    \toprule
         & & \multicolumn{3}{c}{Gender Bias$\downarrow$} & \multicolumn{3}{c}{Recall$\uparrow$} \\
        \cmidrule(lr){3-5}\cmidrule(lr){6-8}
        Dataset & Method & Bias@1 & Bias@5 & Bias@10 & Recall@1 & Recall@5 & Recall@10 \\
    \midrule
        \multirow{4}{*}{COCO1K} & SCAN & .1250 & .2044 & .2506 & 47.7  & $82.0$ & $\mathbf{91.0}$ \\
        & FairSample & .1140 & .1951 & $.2347$ & $\mathbf{49.7}$ & $\mathbf{82.5}$ & 90.9 \\
        & CLIP & .0900 & .2024 & .2648 & $48.2$ & 77.9 & 88.0\\
        & CLIP-clip & $\mathbf{.0670}$ & $\mathbf{.1541}$ & $\mathbf{.2057}$ & 46.1 & 75.2 & 86.0 \\
    \midrule
        \multirow{4}{*}{COCO5K} & SCAN & .1379 & .2133 & .2484 & 25.4 & 54.1 & $67.8$ \\
        & FairSample & .1133 & .1916 & .2288 & $26.8$ & $\mathbf{55.3}$ & $\mathbf{68.5}$ \\
        & CLIP & .0770 & $.1750$ & $.2131$ & $\mathbf{28.7}$ & 53.9 & 64.7\\
        & CLIP-clip & $\mathbf{.0672}$ & $\mathbf{.1474}$ & $\mathbf{.1611}$ & 27.3 & 50.8 & 62.0 \\
    \midrule
        \multirow{4}{*}{Flickr30K} & SCAN & .1098 &  .3341 & .3960 & 41.4 & 69.9 & 79.1 \\
        & FairSample & $\mathbf{.0744}$ & $\mathbf{.2699}$ & .3537 & 35.8 & 67.5 & 77.7 \\
        & CLIP & .1150 & .3150 & .3586 & $\mathbf{67.2}$ & $\mathbf{89.1}$ & $\mathbf{93.6}$ \\
        & CLIP-clip & .0960 & .2746 & $\mathbf{.2951}$ & 63.9 & 85.4 & 91.3\\
    \bottomrule
    \end{tabular}
    \caption{Results on MS-COCO (1K and 5K) and Flickr30K test sets. We compare the baseline models (SCAN~\cite{SCAN} and CLIP~\cite{CLIP}) and our debiasing methods (FairSample and CLIP-clip) on both the gender bias metric Bias@K and the retrieval metric Recall@K.
    }
    \label{tab:main_result}
\end{table*}

\paragraph{Debiasing Effectiveness} 
As shown in \cref{tab:main_result}, both the in-processing sampling strategy \emph{FairSample} and the post-processing feature pruning algorithm \emph{clip} consistently mitigate the gender bias on test data. For instance, among the top-10 search images, SCAN with FairSample reduces gender bias from $0.3960$ to $0.3537$ (decreased by $10.7\%$) on Flickr30K. Using the clipped CLIP features for image search (CLIP-clip), the gender bias drops from $0.2648$ to $0.2057$ ($22.3\%$) on MS-COCO 1K, from $0.2131$ to $0.1611$ ($24.4\%$) on MS-COCO 5K, and from $0.3586$ to $0.2951$ (17.7\%) on Flickr30K. For the tradeoff, CLIP-clip sacrifices the recall performance slightly (from $93.6\%$ Recall@10 to $91.3\%$ on Flickr30K). On the other hand, SCAN with FairSample even achieves a comparable recall performance with SCAN.

\subsection{Gender Bias at Different Top-K Results} 
We plot how gender bias varies across different values of $K$ (1-10) for all the compared methods in \cref{fig:gender-bias}. 
We observe that when $K < 5$, the gender bias has a higher variance due to the inadequate retrieved images. When $K \geq 5$, the curves tend to be flat. This result indicates that $\bias@10$ is more recommended than $\bias@1$ for measuring gender bias as it is more stable. It is also noticeable that CLIP-clip achieves the best fairness performance in terms of $\bias@10$ consistently on all three test sets compared to the other models.

\subsection{Tradeoff between Recall and Bias}\label{sec:trade-off}
There is an inherent tradeoff between fairness and accuracy in fair machine learning~\cite{NEURIPS2019_b4189d9d}. To achieve the best recall-bias tradeoff in our methods, we further examine the effect of the controlling hyper-parameters: the weight $\alpha$ in FairSampling and the number of clipped dimensions $m$ in CLIP-clip. 

\begin{figure}[tb]
    \centering
    \begin{subfigure}[b]{0.48\linewidth}
         \includegraphics[width=\linewidth]{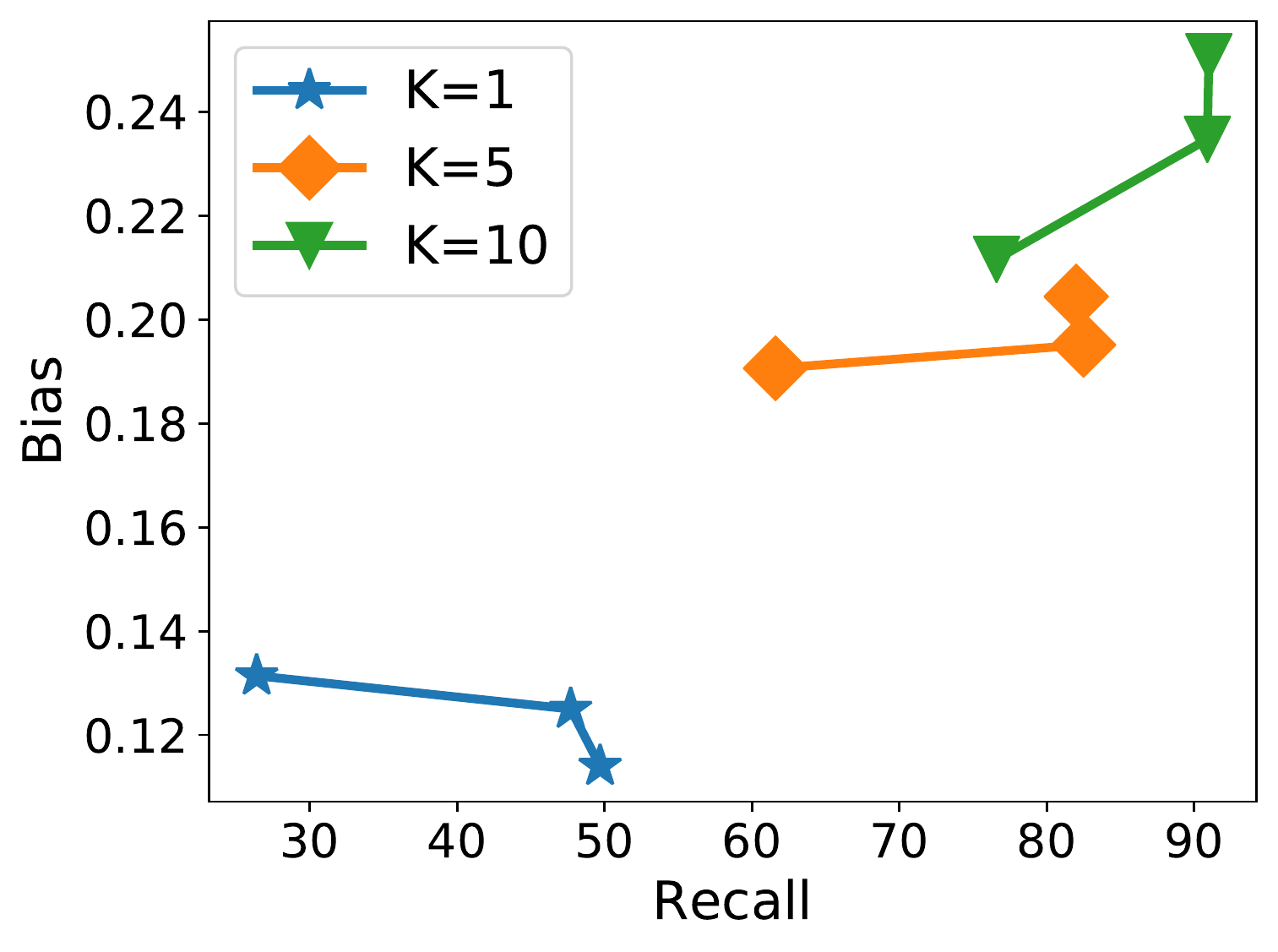}  
         \caption{MS-COCO 1K test set}
    \end{subfigure}
    \hfill
    \begin{subfigure}[b]{0.48\linewidth}
         \includegraphics[width=\linewidth]{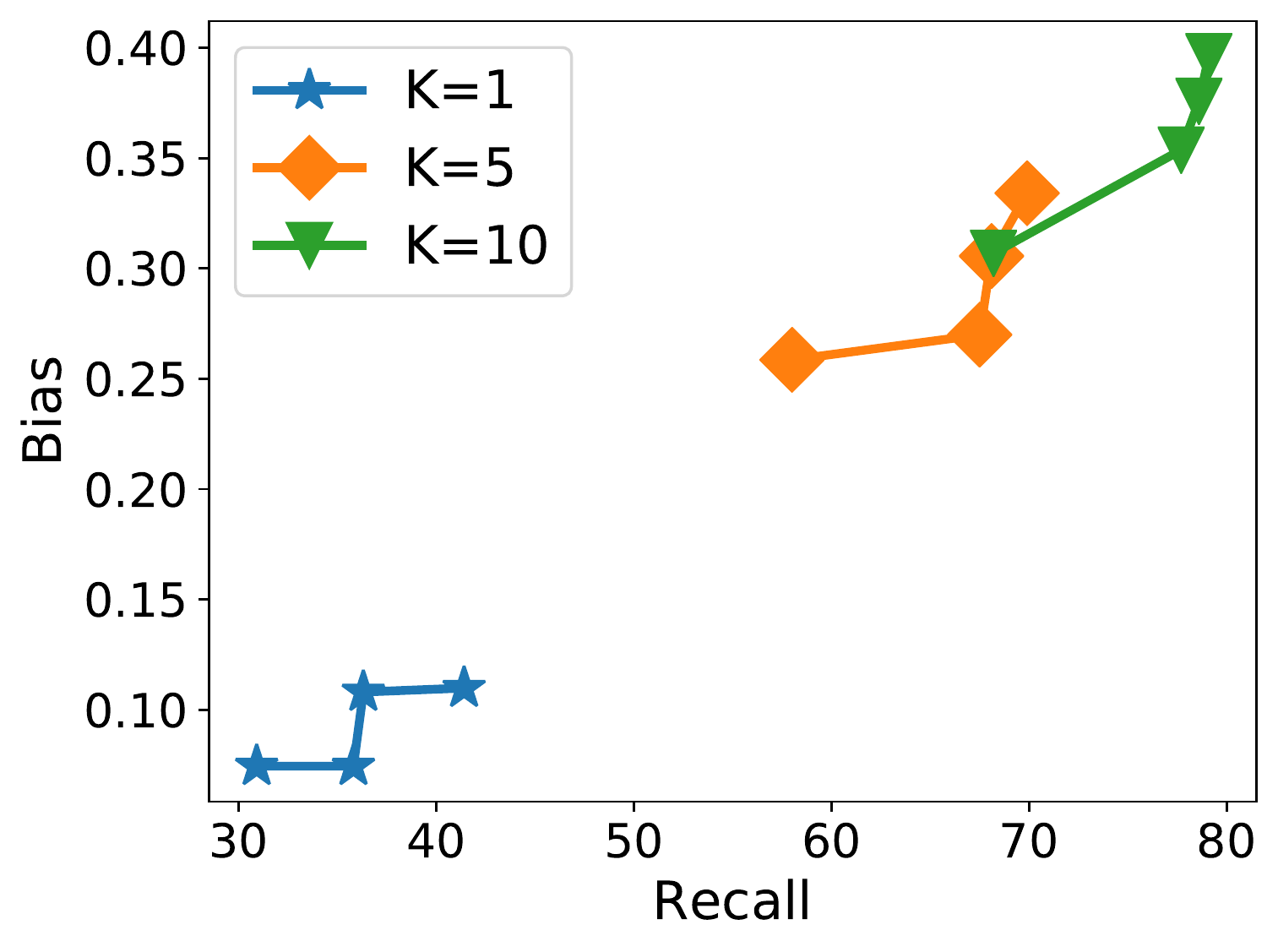}  
         \caption{Flickr30K test set}
    \end{subfigure}
    \vspace{-1ex}
    \caption{The Pareto frontier of recall-bias tradeoff curve for FairSample on MS-COCO 1K and Flickr30K.}
    \label{fig:fairsample_tradeoff}
\end{figure}

\cref{fig:fairsample_tradeoff} demonstrates the recall-bias curve with the fair sampling weight $\alpha \in [0,1]$. Models of higher recall often suffer higher gender bias, but the fairness improvement outweighs the recall performance drop in FairSample models. For example, the model fully trained with fair sampling ($\alpha=1$) has the lowest bias and drops the recall performance the most---it relatively reduces $22.5\%$ Bias@10 but only decreases $10.9\%$ Recall@10 on Flickr30K. 
We choose $\alpha=0.4$ for the final model, which has a better tradeoff in retaining the recall performance.

\begin{figure}[t]
    \centering
    \begin{subfigure}[b]{0.445\linewidth}
    \centering
    \includegraphics[width=\linewidth]{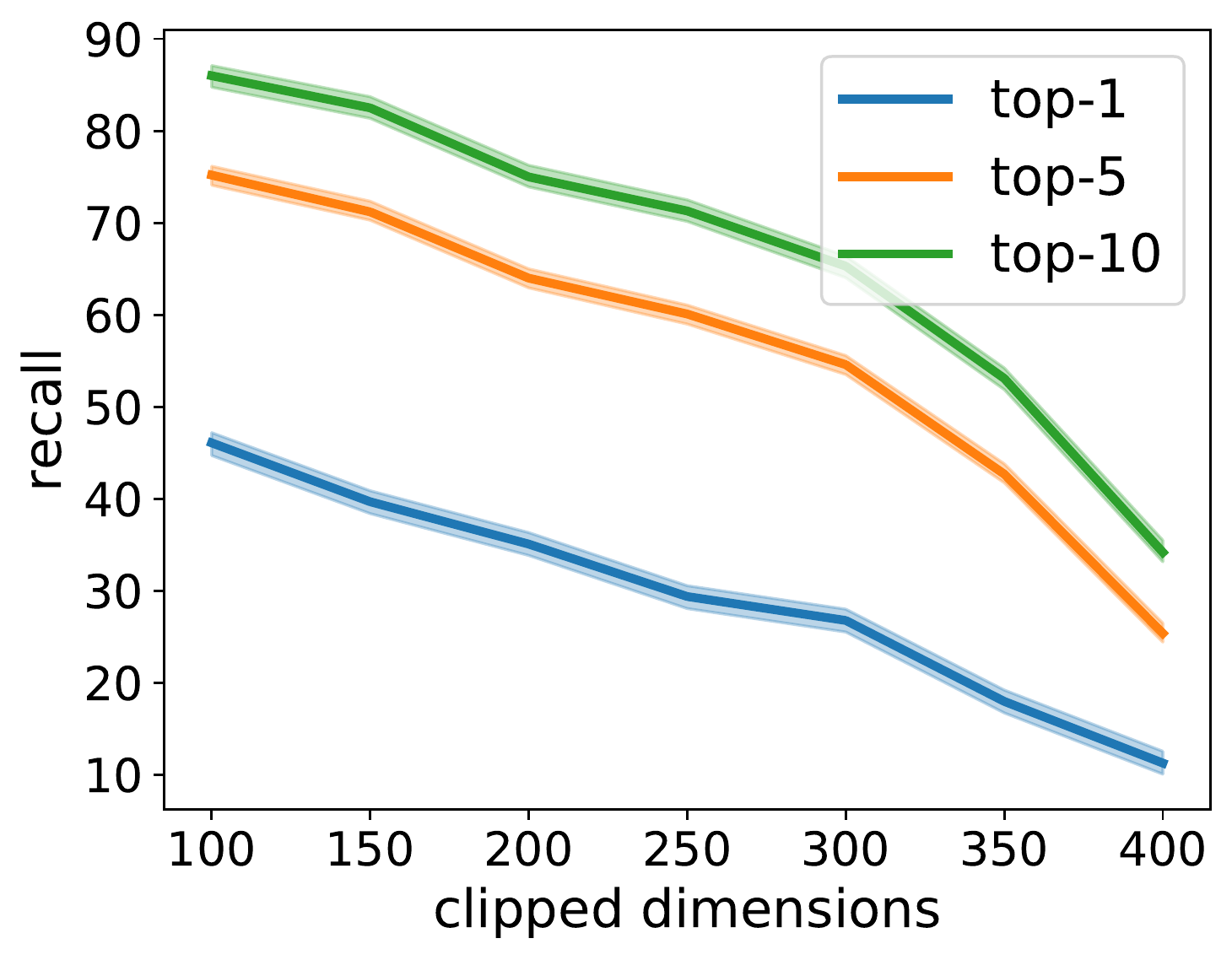}
    \caption{recall}
    \end{subfigure}
    \hfill
    \begin{subfigure}[b]{0.475\linewidth}
    \centering
    \includegraphics[width=\linewidth]{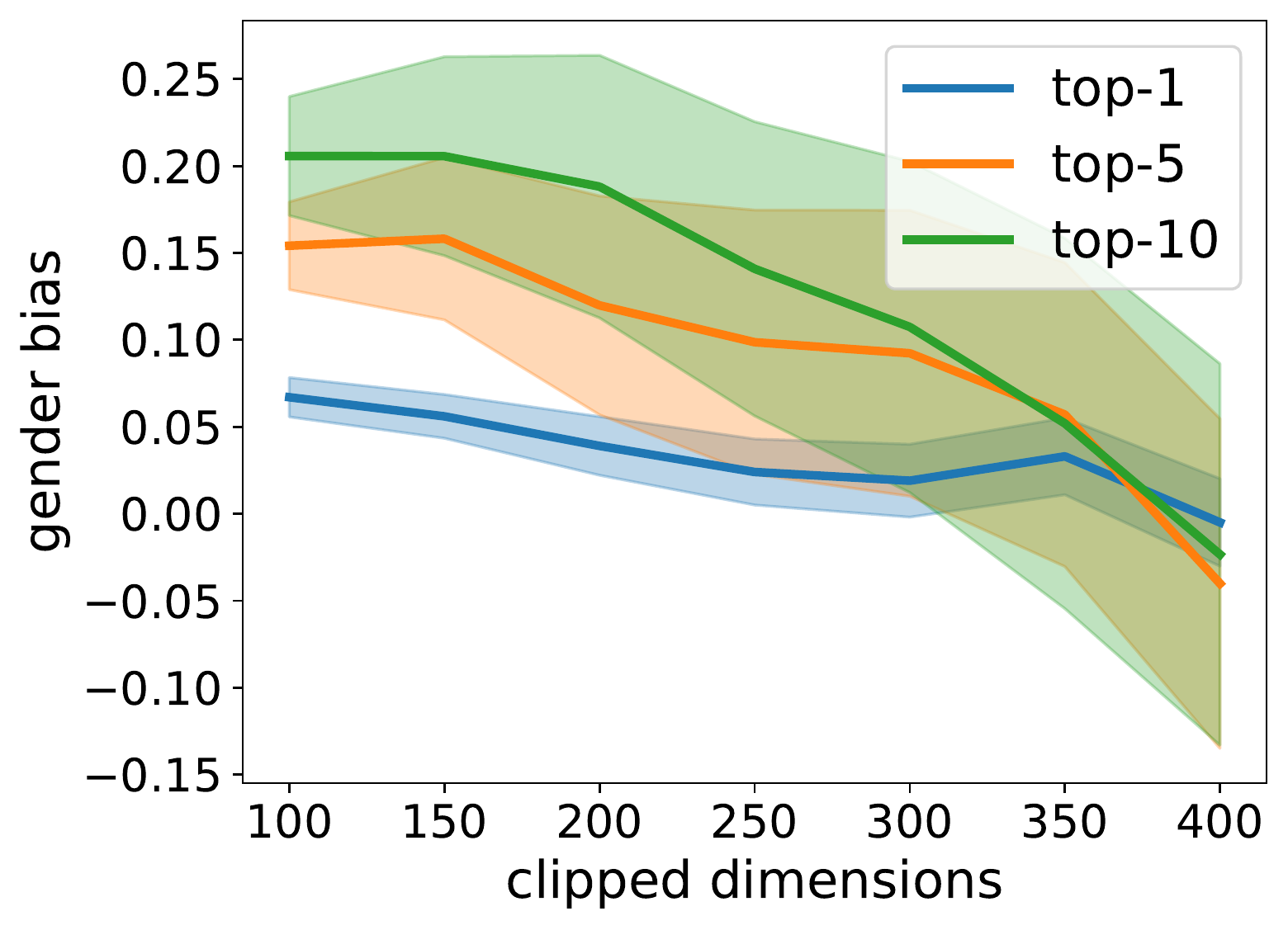}
    \caption{gender bias}
    \end{subfigure}
    \caption{Effect of the number of clipped dimensions $m$ on performance of recall and bias on MS-COCO 1K.}
    \label{fig:clip-ablation}
\end{figure}
As shown in \cref{fig:clip-ablation}, we set the range of the clipping dimension $m$ between $100$ and $400$ on MS-COCO 1K. We find that clipping too many covariates (1) harms the expressiveness of image and text representations (Recall@1 drops from $46.1\%$ to $11.3\%$, Recall@5 drops from $75.2\%$ to $25.4\%$, and Recall@10 drops from $86.0\%$ to $34.2\%$), and (2) causes high standard deviation in gender bias. In light of the harm on expressiveness, we select $m=100$ for conventional use.

\begin{figure*}[htb]
\centering
\begin{subfigure}[b]{0.49\linewidth}
    \centering
    \includegraphics[width=\textwidth]{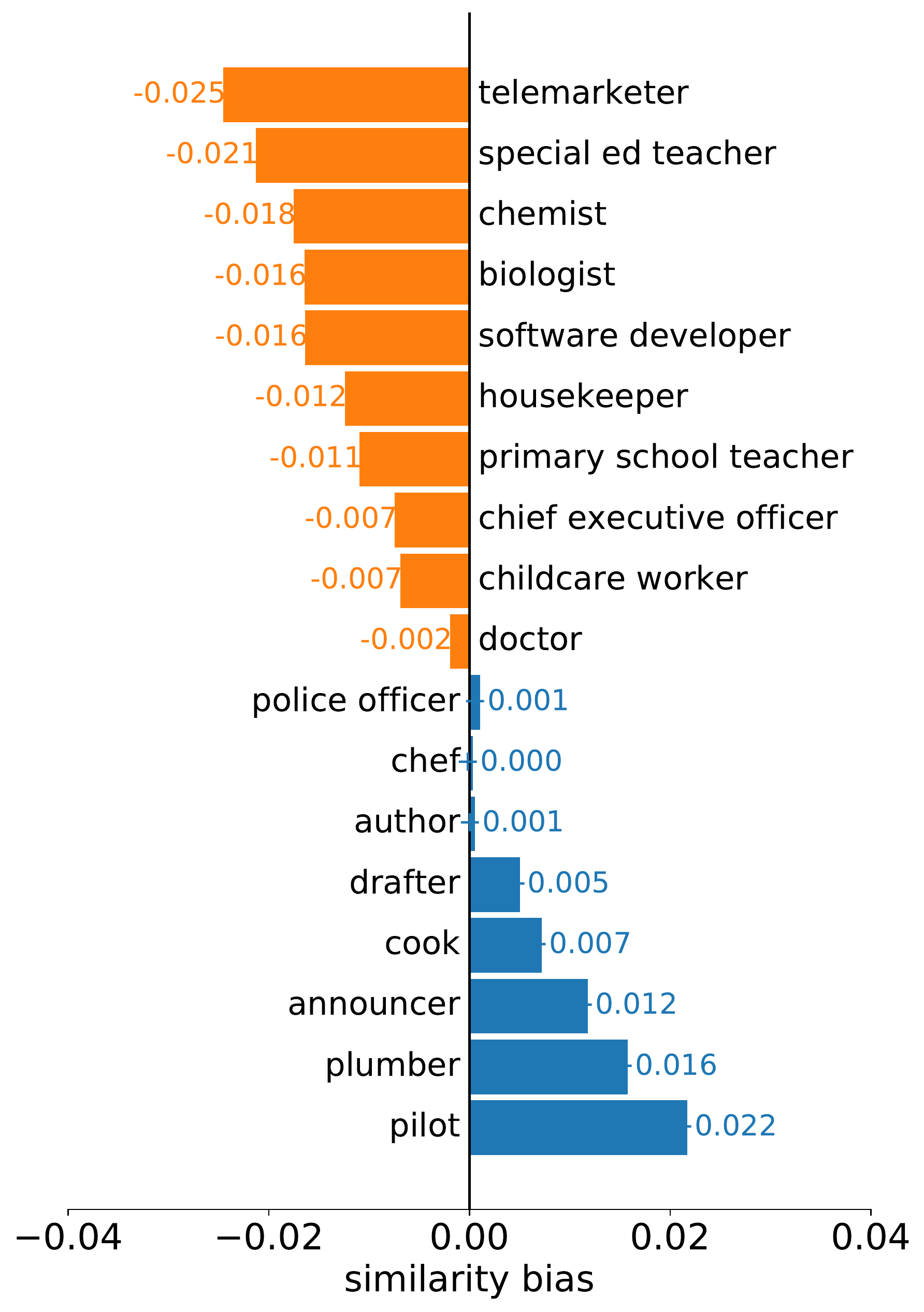}
    \caption{CLIP}
\end{subfigure}
\begin{subfigure}[b]{0.49\linewidth}
    \centering
    \includegraphics[width=\textwidth]{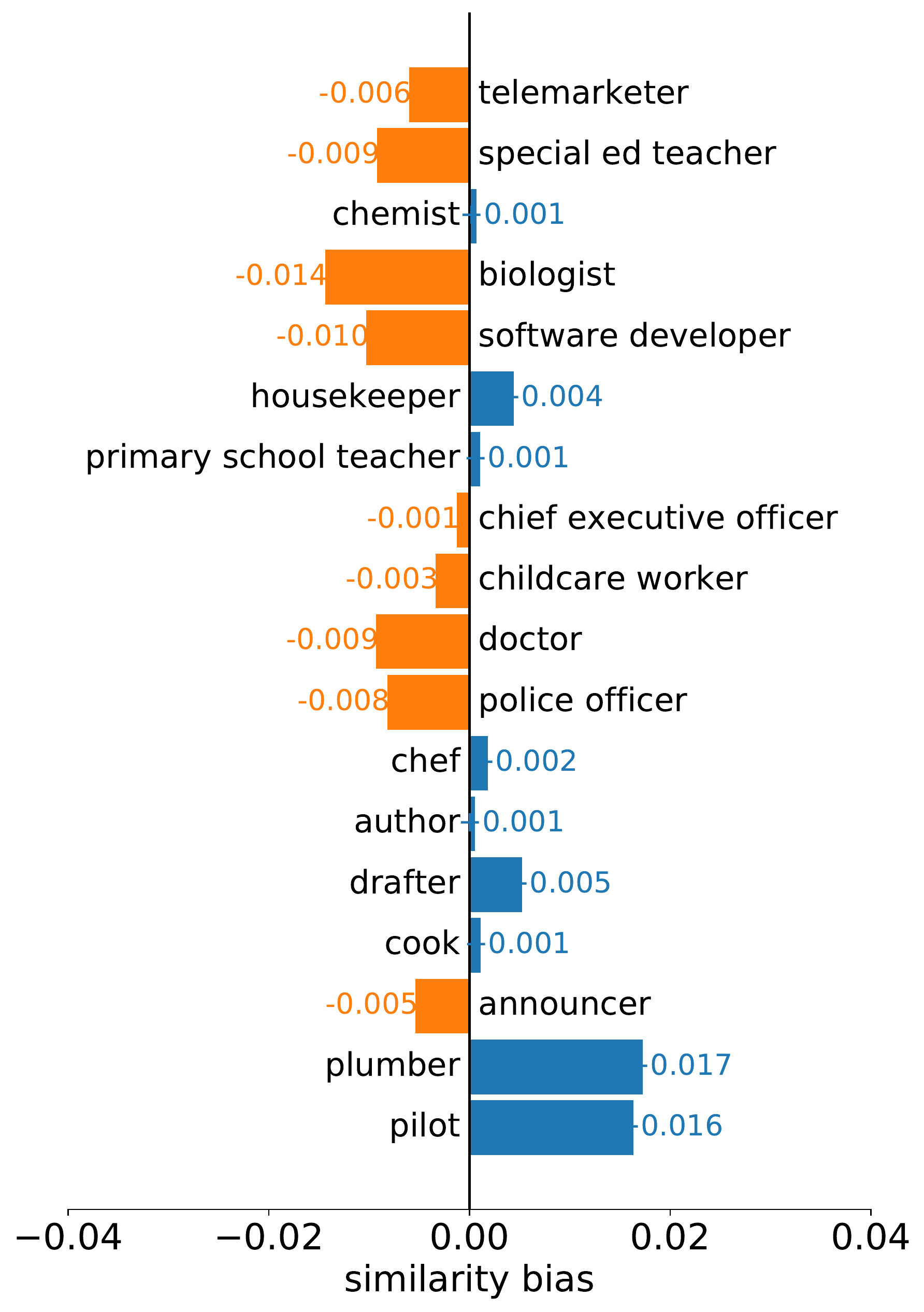}
    \caption{CLIP-clip}
\end{subfigure}
\caption{Gender bias evaluation of internet image search results on \texttt{occupations}~\cite{gender-image-search}. We visualize the similarity biases on 18 occupations. \crule[RoyalBlue]{8pt}{8pt} indicates the occupation is biased towards males and \crule[orange]{8pt}{8pt} indicates it is biased towards females. The clip algorithm mitigates gender bias for a variety of occupations.
}\label{fig:occupation}
\end{figure*}

\subsection{Evaluation on Internet Image Search}
The aforementioned evaluation results on MS-COCO and Flickr30K datasets are limited that they rely on gender labels extracted from human captions. In this sense, it is important to measure the gender biases on a benchmark where the gender labels are identified by crowd annotators. To this end, we further evaluate on the \texttt{occupation} dataset~\cite{gender-image-search}, which collects top 100 Google Image Search results for each gender-neutral occupation search term.\footnote{The data is available at \url{https://github.com/mjskay/gender-in-image-search}.} 
Each image is associated with the crowd-sourced gender attribute of the participant portrayed in the image. Inspired by \citet{Burns2018WomenAS} and \citet{Tang2020MitigatingGB}, we measure the gender bias by computing the difference of expected cosine similarity between male and female occupational images. Given an occupation $o$, the similarity bias is formulated as
\begin{equation}
    \bias = \Expectation_{v \in \mathcal{V}^o_\male} S(v, o) - \Expectation_{v \in \mathcal{V}^o_\female} S(v, o)
\end{equation}
where $\mathcal{V}^o_\male$ and $\mathcal{V}^o_\female$ are the sets of images for occupation $o$, labeled as ``male'' and ``female''. 

\cref{fig:occupation} demonstrates the absolute similarity bias of CLIP and CLIP-clip on the \texttt{occupation} dataset for 18 occupations. We observe that the CLIP model exhibits severe similarity discrepancy for some occupations, including telemarketer, chemist, and housekeeper, while the \emph{clip} algorithm alleviates this problem effectively. Note that for doctor and police officer, the CLIP-clip model exaggerates the similarity discrepancy, but the similarity bias is still less than 0.01. In general, CLIP-clip is effective for mitigating similarity bias and obtains a 42.3\% lower mean absolute bias of the 100 occupations than the CLIP model ($0.0064$ \textit{vs.} $0.0111$).

%% file: sections/related-work.tex
\section{Related Work}\label{sec:related-work}
\paragraph{Fairness in Machine Learning} 
A number of unfair treatments by machine learning models were reported recently \cite{angwin2016machine,pmlr-v81-buolamwini18a,10.5555/3157382.3157584,10.1145/3025453.3025727}, and the literature has observed a growing demand and interests in proposing defenses, including regularizing disparate impact~\cite{Zafar2015LearningFC} and disparate treatment~\cite{Hardt2016EqualityOO}, promoting fairness through causal inference~\cite{NIPS2017_6995}, and adding fairness guarantees in recommendations and information retrieval~\cite{Beutel2019FairnessIR,Biega2018EquityOA,10.1145/3397271.3401100}. The existing fair machine learning solutions can be broadly categorized as pre-processing~\cite{KamiranFaisal2012DataPT,Feldman2015,NIPS2017_6988}, in-processing, and post-processing approaches. Pre-processing algorithms typically re-weight and repair the training data which captures label bias or historical discrimination~\cite{KamiranFaisal2012DataPT,Feldman2015,NIPS2017_6988}. %
In-processing algorithms focus on modifying the training objective with additional fairness constraints or regularization terms~\cite{Zafar2017FairnessCM,fairlearn,Cotter2019OptimizationWN}. Post-processing algorithms enforce fairness constraints by applying a post hoc correction of a (pre-)trained classifier~\cite{Hardt2016EqualityOO,NIPS2017_6988}. In this work, the \textit{fair sampling} strategy designed for the contrastive learning framework could be considered as an in-processing treatment, while the \textit{clip} algorithm is in the post-processing regime that features an information-theoretical clipping procedure. Our contribution highlights new challenges of reducing gender bias in a multimodal task and specializes new in-processing and post-processing ideas in the domain of image search.

\paragraph{Social Bias in Multi-modality} 
Implicit social bias related to gender and race has been discussed in multimodal tasks including image captioning~\cite{Burns2018WomenAS,Tang2020MitigatingGB}, visual question answering~\cite{Manjunatha_2019_CVPR}, face recognition~\cite{pmlr-v81-buolamwini18a}, and unsupervised image representation learning~\cite{unsupervised-image-representation}. For example, \citet{Zhao2017MenAL} shows that models trained on unbalanced data can amplify bias, and injecting corpus-level Lagrangian constraints can calibrate the bias amplification. \citet{Caliskan2017SemanticsDA} demonstrates the association between the word embeddings of occupation and gendered concepts correlates with the imbalanced distribution of gender in text corpora. There are also a series of debiasing techniques in this area. \citet{10.5555/3157382.3157584} propose to surgically alter the embedding space by identifying the gender subspace from gendered word pairs. \citet{manzini-etal-2019-black} extend the bias component removal approach to the setting where the sensitive attribute is non-binary. Data augmentation approaches remove the implicit bias in the training corpora and train the models on the balanced datasets~\cite{zhao-etal-2018-gender}. Our work complements this line of research by examining gender bias induced by multimodal models in image search results. Our focus on gender bias in the gender-neutral language would offer new insights for a less explored topic to the community.

\paragraph{Gender Bias in Online Search Systems} 
Our work is also closely connected to studies in the HCI community showing the gender inequality in online image search results. \citet{gender-image-search} articulate the gender bias in occupational image search results affect people's perceptions of the prevalence of men and women in each occupation. \citet{gender-image-search} compare gender proportions in occupational image search results and discuss how the bias affects people's perceptions of the prevalence of men and women in each occupation. \citet{DBLP:journals/jasis/SinghCIF20} examine the prevalence of gender stereotypes on various digital media platforms. \citet{10.1145/3025453.3025727} identify gender bias with character traits. Nonetheless, these works do not attempt to mitigate gender bias in search algorithms. Our work extends these studies into understanding how gender biases enter search algorithms and provides novel solutions to mitigating gender bias in two typical model families for image search. 

%% file: sections/conclusion.tex
\section{Conclusion}
In this paper, we examine gender bias in image search models when search queries are gender-neutral. As an initial attempt to study this critical problem, we formally identify and quantify gender bias in image search. To mitigate the gender bias perpetuating two representative fashions of image search models, we propose two novel debiasing algorithms in in-processing and post-processing manners. When training a new image search model, the in-processing \textit{FairSample} method can be used to learn a fairer model from scratch. Meanwhile, the \textit{clip} algorithm can be used for lightweight deployment of pre-trained representation models with accessible gender information.

%% file: sections/appendix.tex
\section{Gender Word Lists}\label{app:word-list}
We show the word lists for identifying the gender attributes of a caption in \cref{tab:gender-word-list}.
\begin{table}[!ht]
    \centering
    \begin{tabular}{p{0.2\linewidth} | p{0.7\linewidth}}
    \toprule
        feminine words & woman, women, female, girl, lady, mother, mom, sister, daughter, wife, girlfriend  \\
    \midrule
        masculine words & man, men, male, boy, gentleman, father, brother, son, husband, boyfriend \\
    \midrule
        gender-neutral words & person, people, human, adult, baby, child, kid, children, guy, teenage, crowd\\
    \bottomrule
    \end{tabular}
    \caption{Gender word lists. We identify the gender attributes of captions based on the occurrence of gender-specific words appeared in the sentences.}
    \label{tab:gender-word-list}
\end{table}

\section{Implementation Details}
\subsection{Computing Infrastructure}
We use a GPU server with 4 NVIDIA RTX 2080 Ti GPUs for training and evaluation.

\subsection{Computational Time Costs}
We find that SCAN~\cite{SCAN} and SCAN with fair sampling need about 20 hours for training 30 epochs on MS-COCO and 8-10 minutes for testing on 1K test set. In comparison, pre-trained CLIP~\cite{CLIP} and CLIP-clip can be evaluated within 1 minutes on MS-COCO 1K test set.

\section{Qualitative Examples}\label{app:qualitative}

\begin{figure*}[!b]
    \setlength{\tabcolsep}{1ex}
    \renewcommand{\arraystretch}{1.5}
    \centering
    \begin{tabular}{r c c c c c}
        \rotatebox[origin=l]{90}{SCAN} &
        \includegraphics[width=0.16\linewidth, height=0.16\linewidth]{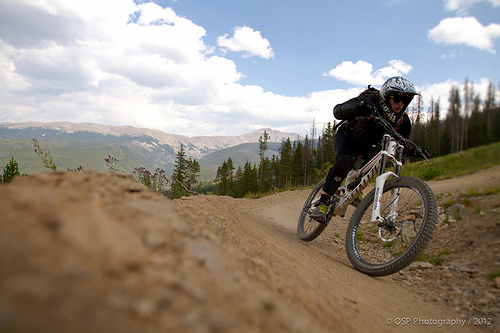} &
        \includegraphics[width=0.16\linewidth,height=0.16\linewidth]{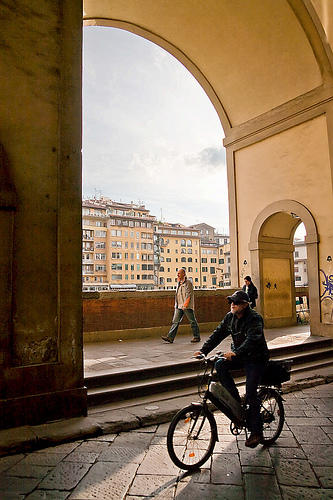} &
        \includegraphics[width=0.16\linewidth,height=0.16\linewidth]{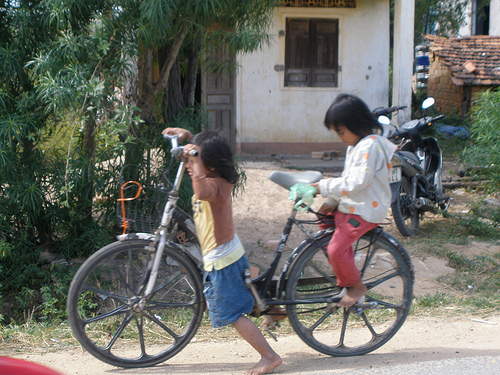} &
        \includegraphics[width=0.16\linewidth,height=0.16\linewidth]{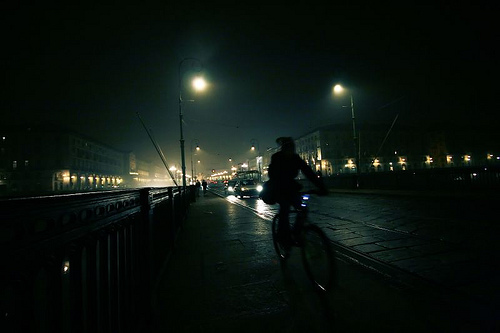} &
        \includegraphics[width=0.16\linewidth,height=0.16\linewidth]{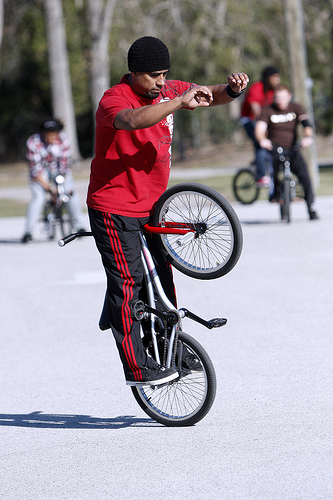} \\
        \rotatebox[origin=l]{90}{FairSample} &
        \includegraphics[width=0.16\linewidth, height=0.16\linewidth]{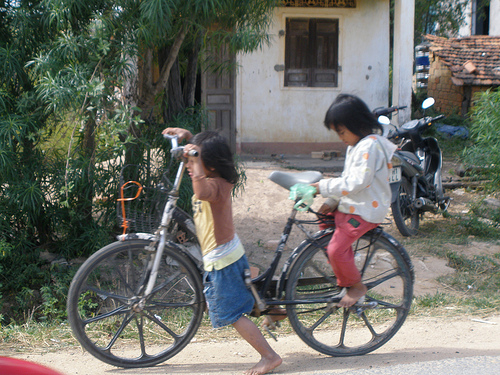} &
        \includegraphics[width=0.16\linewidth,height=0.16\linewidth]{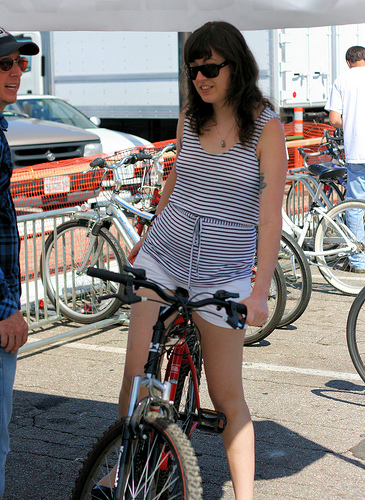} &
        \includegraphics[width=0.16\linewidth,height=0.16\linewidth]{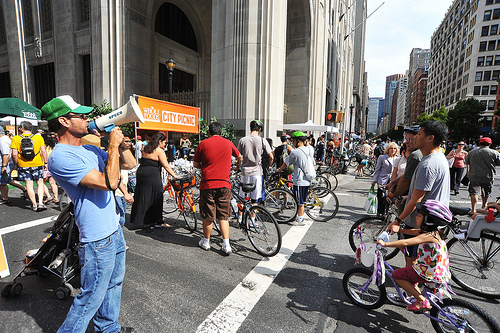} &
        \includegraphics[width=0.16\linewidth,height=0.16\linewidth]{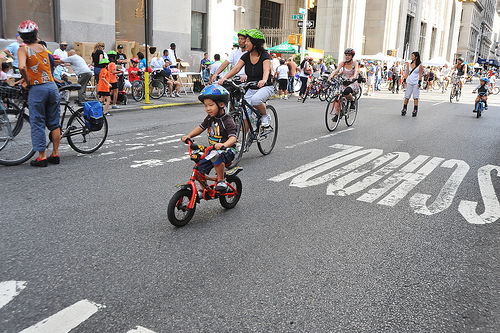} &
        \includegraphics[width=0.16\linewidth,height=0.16\linewidth]{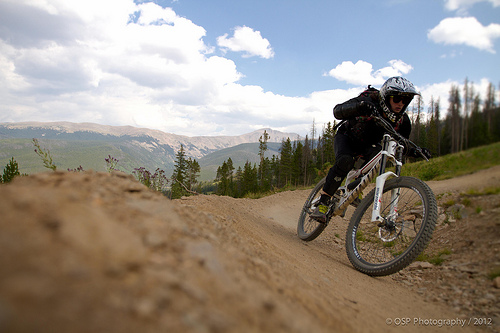} \\
        \rotatebox[origin=l]{90}{CLIP} &
        \includegraphics[width=0.16\linewidth,height=0.16\linewidth]{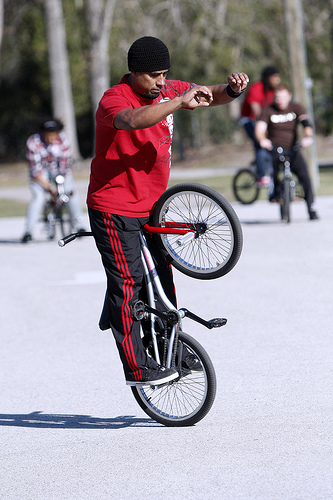} &
        \includegraphics[width=0.16\linewidth,height=0.16\linewidth]{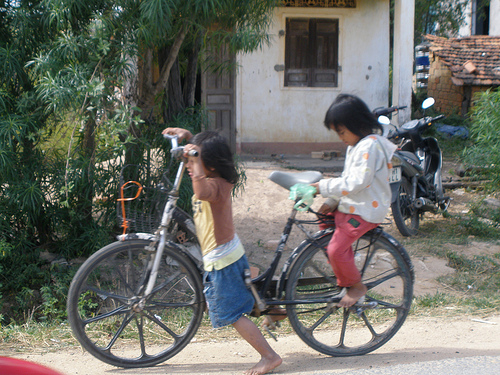} &
        \includegraphics[width=0.16\linewidth,height=0.16\linewidth]{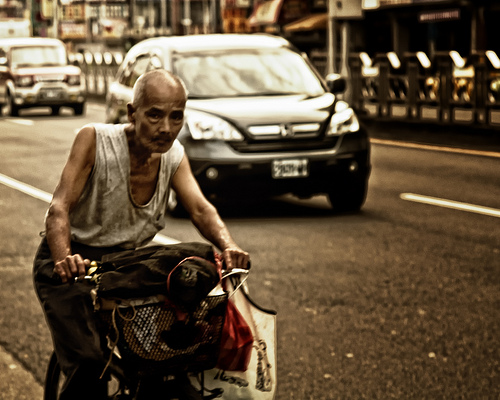} &
        \includegraphics[width=0.16\linewidth,height=0.16\linewidth]{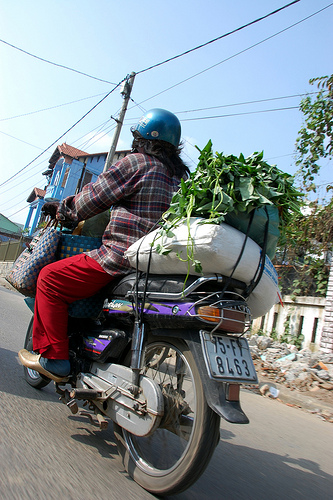} &
        \includegraphics[width=0.16\linewidth, height=0.16\linewidth]{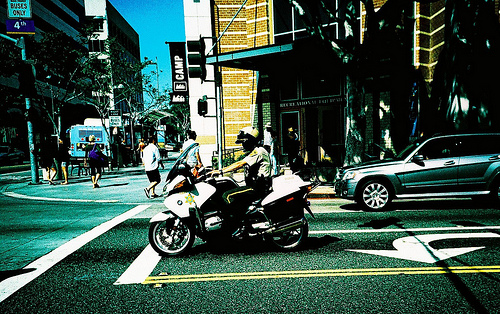} \\
        \rotatebox[origin=l]{90}{CLIP-clip} &
        \includegraphics[width=0.16\linewidth,height=0.16\linewidth]{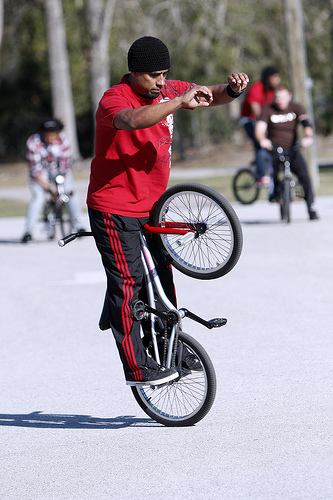} &
        \includegraphics[width=0.16\linewidth,height=0.16\linewidth]{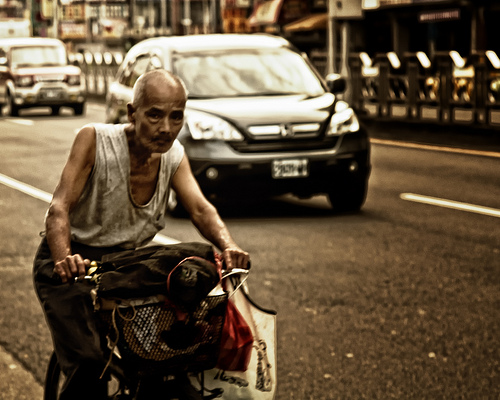} &
        \includegraphics[width=0.16\linewidth,height=0.16\linewidth]{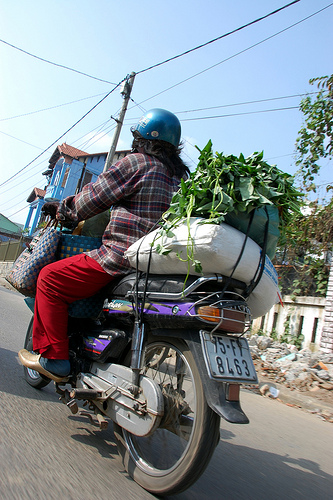} &
        \includegraphics[width=0.16\linewidth,height=0.16\linewidth]{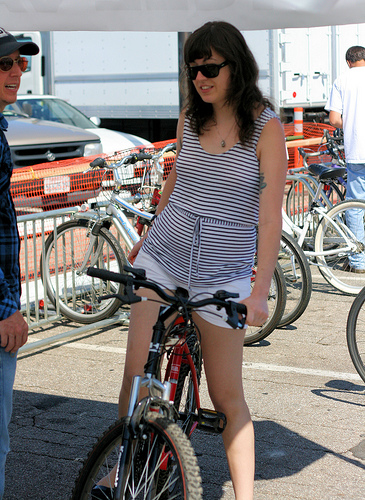} &
        \includegraphics[width=0.16\linewidth, height=0.16\linewidth]{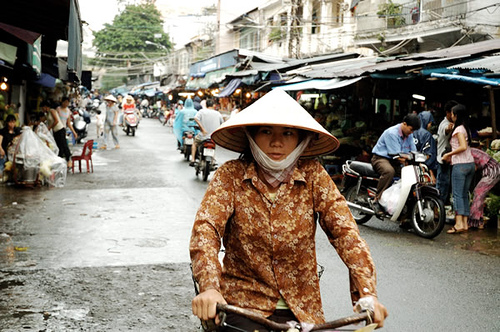} \\
    \end{tabular}
    \caption{Qualitative analysis of gender bias in image search results. The text query is ``a person riding a bike''. The first row presents the top-5 retrieved images for SCAN, the second row presents the top-5 retrieved images for SCAN+FairSample, the third row presents the top-5 retrieved images for CLIP, and the last row presents the top-5 retrieved images for CLIP-clip.
    }
    \label{fig:qualitative-analysis2}
\end{figure*}

We take a qualitative study on the image search results. We show the results of searching ``a person riding a bike'' in \cref{fig:qualitative-analysis2}. The first row presents the top-5 retrieved images for SCAN, the second row presents the top-5 retrieved images for SCAN+FairSample, the third row presents the top-5 retrieved images for CLIP, and the last row presents the top-5 retrieved images for CLIP-clip. While we notice that all the models retrieve relevant images, we find FairSample put images depicting females in a higher rank.

